\pdfoutput=1

\documentclass[11pt]{article}

\usepackage[]{acl}

\usepackage{times}
\usepackage{latexsym}
\usepackage{booktabs}

\usepackage[T1]{fontenc}

\usepackage[utf8]{inputenc}

\usepackage{microtype}

\usepackage{inconsolata}

\usepackage{graphicx}
\usepackage[frozencache,cachedir=minted-cache]{minted}

\usepackage{graphicx}
\usepackage{microtype}
\usepackage{amsmath}
\usepackage{amssymb}
\usepackage{mathtools}
\usepackage{amsthm}
\usepackage{bbm}
\usepackage{subcaption}
\usepackage{multirow}
\usepackage{algpseudocode}
\usepackage[ruled,linesnumbered]{algorithm2e}
\usepackage{enumitem}
\usepackage{booktabs}
\usepackage{soul}

\DeclareRobustCommand{\hlcyan}[1]{{\sethlcolor{cyan}\hl{#1}}}
\DeclareRobustCommand{\hllime}[1]{{\sethlcolor{lime}\hl{#1}}}
\DeclareRobustCommand{\hlpink}[1]{{\sethlcolor{pink}\hl{#1}}}

\title{Re-Invoke: Tool Invocation Rewriting for Zero-Shot Tool Retrieval}

\author{Yanfei Chen$^{1}$\thanks{\ Correspondence to:  \{yanfeichen, jinsungyoon, chenyulee\}@google.com}, \
Jinsung Yoon$^{1}$, \
Devendra Singh Sachan$^{2}$, \\
\bf
Qingze Wang$^{3}$, \
Vincent Cohen-Addad$^{3}$, \
Mohammadhossein Bateni$^{3}$, \\
\bf
Chen-Yu Lee$^{1}$, \
Tomas Pfister$^{1}$ \\
$^1$Google Cloud AI Research,
$^2$Google DeepMind,
$^3$Google\\
}

\begin{document}
\maketitle
\begin{abstract}
Recent advances in large language models (LLMs) have enabled autonomous agents with complex reasoning and task-fulfillment capabilities using a wide range of tools. 
However, effectively identifying the most relevant tools for a given task becomes a key bottleneck as the toolset size grows, hindering reliable tool utilization.
To address this, we introduce Re-Invoke, an unsupervised tool retrieval method designed to scale effectively to large toolsets without training.
Specifically, we first generate a diverse set of synthetic queries that comprehensively cover different aspects of the query space associated with each tool document during the tool indexing phase. 
Second, we leverage LLM's query understanding capabilities to extract key tool-related context and underlying intents from user queries during the inference phase. 
Finally, we employ a novel multi-view similarity ranking strategy based on intents to pinpoint the most relevant tools for each query.
Our evaluation demonstrates that Re-Invoke significantly outperforms state-of-the-art alternatives in both single-tool and multi-tool scenarios, all within a fully unsupervised setting. 
Notably, on the ToolE datasets, we achieve a 20\% relative improvement in nDCG@5 for single-tool retrieval and a 39\% improvement for multi-tool retrieval.
\end{abstract}

\section{Introduction}
Recently, large language models (LLMs) have demonstrated impressive capabilities on a variety of complex tasks, including math, reasoning and coding \citep{openai2023gpt4, anil2023palm, gemini2023}.
They can even surpass average human performance on standardized exams such as college entrance tests, law school admission, and math competitions \citep{zhong2023agieval}.
However, LLMs are pre-trained on a static corpus, limiting their adaptability to the rapidly evolving real world, and frequent fine-tuning \citep{wei2021finetuned} is computationally expensive.

In contrast, humans leverage a vast array of tools to interact with the external world, using search engines for information retrieval, maps for navigation, calculators for algebraic tasks, and so on. 
Augmenting LLMs with external tools, rather than relying solely on their internal knowledge, could unlock their potential to tackle even more challenging problems. 
This insight has driven recent interests in both academic research \citep{parisi2022talm, schick2023toolformer, lu2023chameleon, cai2023large, patil2023gorilla, hsieh2023tool, qin2023toolllm} and industrial applications. 
Examples include ChatGPT plugins \citep{chatgptplugins}, with supported third-party APIs, and Bard extensions \citep{bardextensions}, which connect to Google APIs and services.

\begin{figure}[t!]
    \centering
    \includegraphics[width=0.49\textwidth]{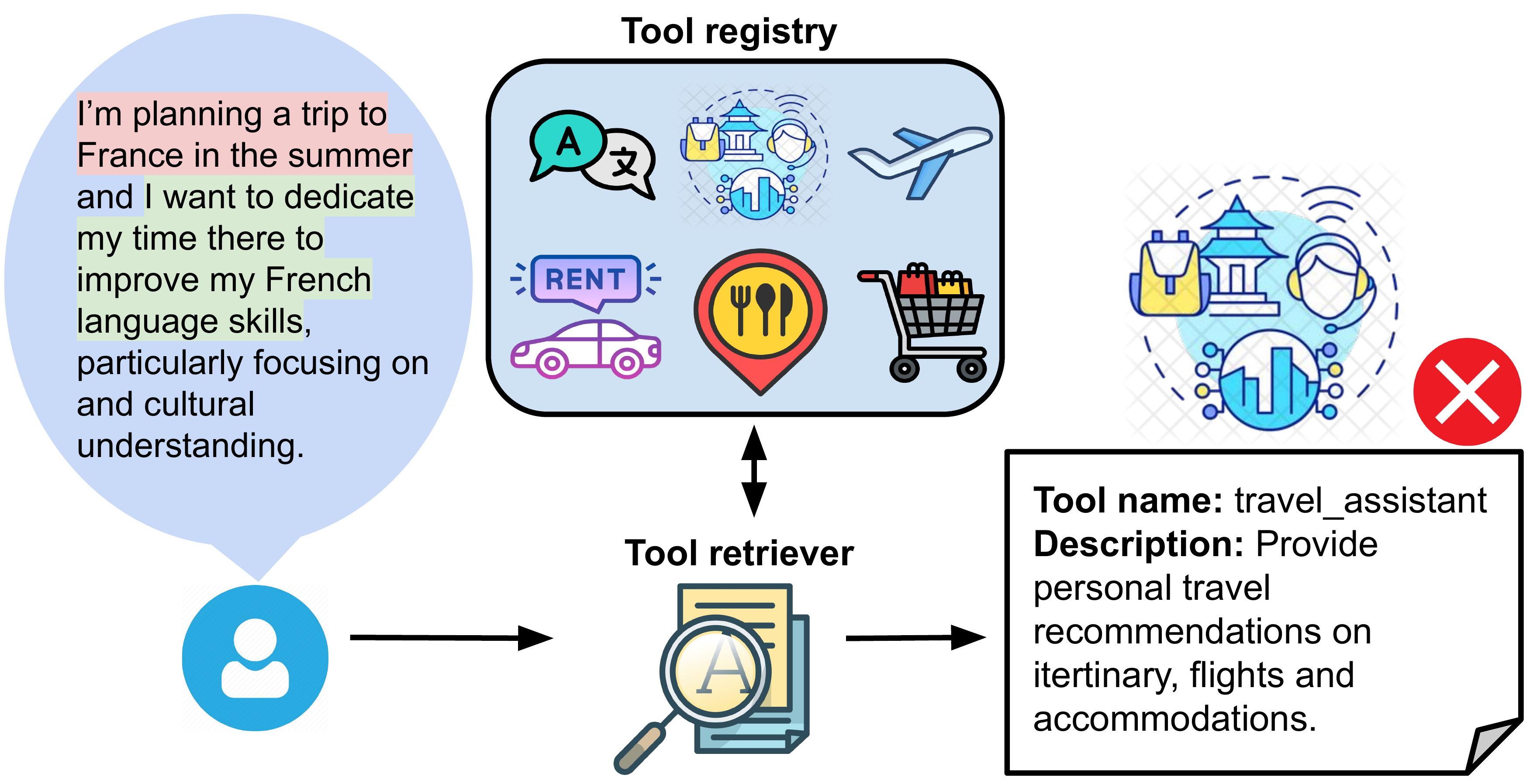}
    \caption{An example of low-performance retrieval methods failing to identify the actual user intents \hllime{``improve French language skills''} due to the context \hlpink{``planning a trip to France''}. It selects similar, but incorrect travel assistant tool instead of the ground-truth language learning tool from the given pool of tools.
    }
    \label{fig:intro}
\end{figure}

Common approaches to integrate tools with LLMs often rely on supervised methods to generate tool calling functions \citep{schick2023toolformer, patil2023gorilla, parisi2022talm, qin2023toolllm, hao2024toolkengpt} or in-context learning by providing the tool documents and few-shot demonstrations \citep{xu2023tool, lu2023chameleon, hsieh2023tool}.
However, these methods face practical challenges when scaling to a large number of tools: 
(a) \textit{Input Token Length Limitations}: LLMs have inherent input token length limitations, making it infeasible to include a comprehensive list of tools within a single prompt. 
Moreover, LLMs can struggle to effectively process relevant information from lengthy input contexts \citep{liu2023lost}. 
(b) \textit{Evolving Tool Pool}: LLMs are often paired with a tool retriever trained on labeled query-tool pairs. However, the ideal LLM toolkit should be vast and dynamic, with tools undergoing frequent updates.
Providing and maintaining labels for such an extensive and evolving toolset is impractical. 
Continuous retraining would also require extensive production maintenance. (c) \textit{Ambiguous User Intents}: User contexts in the queries could obfuscate the underlying intents and failure to identify the intents could lead to calling the wrong tools (See Fig. \ref{fig:intro}).

To address these unique challenges, we introduce \textbf{Re-Invoke}, a novel unsupervised retrieval method designed for tool uses. 
Re-Invoke leverages LLMs for both tool document enrichment and user intent extraction, thereby enhancing tool retrieval performance across various use cases. 
Specifically, we propose two key components to guide unsupervised tool retrieval: 
(a) a \textit{query generator} to automatically enrich the tool document with diverse synthetic queries during offline indexing, 
(b) an \textit{intent extractor} to extract the underlying tool-related intents from verbose user queries during online inference. 
Our approach consistently and significantly improves upon state-of-the-art alternatives, achieving a 20\% relative improvement in nDCG@5 on single-tool retrieval tasks and 39\% improvement on multi-tool retrieval tasks with ToolE dataset.

\section{Related Work}
\paragraph{Tool Retrievals for Tool-Use.}
ReAct \citep{yao2022react} pioneers the interaction and reasoning with diverse tools using in-context reasoning traces, particularly in decision-making and multi-step reasoning environments. 
\citet{schick2023toolformer} proposes a self-supervised training method with API demonstrations.
\citet{patil2023gorilla} and \citet{hsieh2023tool} demonstrate that augmenting LLMs with tool documentation significantly improves their ability to generate correct API calls by mitigating hallucinations, compared to prompting with demonstrations alone.
\citet{yuan2024easytool} also shows unifying tool instruction leads to better tool usage.
However, tool document retrieval for LLM tool learning is currently under-explored, as most work simply uses LLM agents to retrieve a limited number of tools.
\citet{patil2023gorilla} first demonstrate that LLMs generate more reliable outputs with the integration of a retrieval system using BM25 \citep{robertson2009probabilistic} and GPT-index \citep{liulamaindex2022}, but still introduce more hallucination and errors compared to the ground truth retriever.
Some works \citep{qin2023toolllm, kong2023tptu, gao2024confucius} train a Sentence-Bert transformer model using the fully labeled query-API documentation pairs as a tool retriever. 
The key distinction between our approach and existing tool retrieval systems lies in our emphasis on zero-shot usage, eliminating the need for any labeled data.

\paragraph{Generative Document Expansion.}
Appending relevant terms, such as queries, to documents effectively enriches document representation for sparse retrievals. 
\citet{nogueira2019document} demonstrated this by using a language model to generate search queries for improved retrieval in search engines. 
\citet{lewis2021paq} introduced Probably Asked Questions (PAQ) by generating the question given a passage and an answer, and a retriever trained using PAQ demonstrated the strength in accuracy, speed, and space efficiency for selective QA. 
\citet{ma2023pre} trains a dense retriever after applying document expansion. 
Our approach also leverages document expansion with generative language models, but with a focus on tool selection rather than search engine or question-answering tasks.
We further emphasize the ability to extract the user intents from queries to better match the varying complexities of downstream tasks.

\begin{figure*}[ht]
\centering
\includegraphics[width=0.98\textwidth]{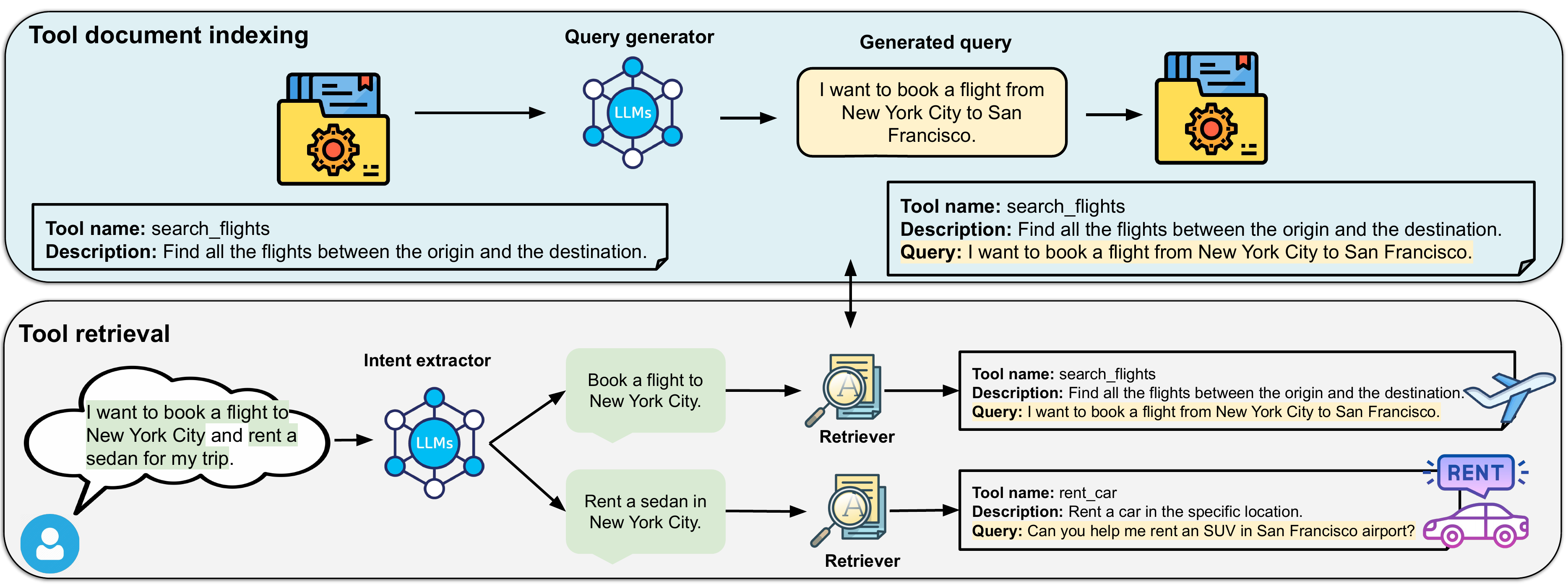}
\caption{An overview of Re-Invoke for tool retrieval tasks. 
(\textbf{Top}) A query generator generates diverse synthetic queries from the tool documents and each synthetic query is concatenated with the tool document to create multiple copies of the expanded tool document. 
(\textbf{Bottom}) An intent extractor synthesizes multiple underlying intents from the user queries in order to retrieve the relevant tools.}
\label{fig:workflow}
\vspace{2mm}
\end{figure*}

\paragraph{Generative Query Expansion.}
Augmenting user queries with hypothetical information is a popular approach in both dense and sparse retrieval methods. 
Query2doc \citep{wang2023query2doc} expands queries with pseudo-generated documents through few-shot prompting.
\citet{jagerman2023query} further extends this idea by studying different prompting methods.
\citet{liu2022query} improves query expansion by balancing diversity and relevance through a combination of effective filtering and documents fusion. 
\citet{shen2023large} augments queries with potential answers by prompting LLMs with a composition of the query and its in-domain candidates. 
\citet{mackie2023generative} enriches the original query with useful terms from diverse generation subtasks.
\citet{chuang2023expand} proposes a query expansion and reranking approach to train a re-ranker after query expansion. 
Alternatively, \citet{gao2022precise} proposes a zero-shot dense retrieval system by first instructing LLMs to generate a hypothetical document given the query for semantic retrievals.
Those approaches primarily focus on generative pseudo-relevance feedback by enriching user queries within the retrieval system. 
They are fundamentally different from our approach, which focuses on query understanding rather than query expansion.

\paragraph{Query Rewriting.}
LLM-aided query rewriting is commonly used in conversational search engine to precisely understand user's contextual search intent through in-context learning \citep{yu2020few, ye2023enhancing, mao2023large, anand2023context}. 
Some works even train the query rewriter in a rewrite-retrieve-read pipeline, allowing interaction with the search engine \citep{feng2023knowledge, ma2023query}.
While LLMs are primarily used to summarize user context in conversations in these works, Re-Invoke focuses on extracting underlying intents for tool uses, rather than solely for information retrieval.

\section{Method: Re-Invoke}
We formulate the tool retrieval task as retrieving the most relevant tools that a downstream agent can execute to fulfill user queries, given a list of tool documents describing the intended tool usage.

Re-Invoke, our proposed fully unsupervised retrieval method designed for tool retrieval tasks, is illustrated in Fig.~\ref{fig:workflow}. 
It consists of two core components: 
(1) \textit{Query generator}: for each tool document, LLMs generate diverse synthetic queries answerable by the corresponding tool. 
These queries enrich the tool document and are then indexed by encoding them into the embedding space when the tool documents are ingested offline.
(2) \textit{Query intent extractor}: during online inference, LLMs extract the core tool-related request(s) from user queries, filtering out irrelevant background context.
Each user intent is then encoded into the same embedding space as the tool documents for similarity matching. 
Pseudo-code of Re-Invoke is described in Algorithm~\ref{algo:reinvoke}.

\begin{algorithm}[!ht]
\caption{Pseudo-code of Re-Invoke} \label{algo:reinvoke}
\KwData{Query $Q$, List of tool documents $D$, The number of tools to retrieve $k$, Large language model $\mathcal{L}$, Text embedding model $f_\text{enc}$, $m$ number of synthetic queries generated per each tool document}
\KwResult{$k$ retrieved tool documents $\hat{D}_{1,...,k}$}

\SetKwFunction{FReInvoke}{ReInvoke}
\SetKwProg{Fn}{Function}{:}{\KwRet $\hat{D}_{1,...,k}$}
\Fn{\FReInvoke{$q, D, k, \mathcal{L}, f_\text{enc}$}}{
\For{$d \in D$}{
    \For{$i=1, ..., m$}{
        $d_{i} \gets \texttt{Concat}(d, \mathcal{L}(d)$)\;
    }
}
$q_{1}, ..., q_{n} \gets \mathcal{L}(Q)$\;
$\hat{s}(Q, D) \gets \texttt{rank}(q_{1...n}, d_{1...m}, f_{\text{enc}})$\;
$\hat{D}_{1,...,k} = \texttt{Argmax}(\hat{s}(Q, D), k) $\;
}
\end{algorithm}

\begin{figure*}[ht]
\centering
\includegraphics[width=\textwidth]{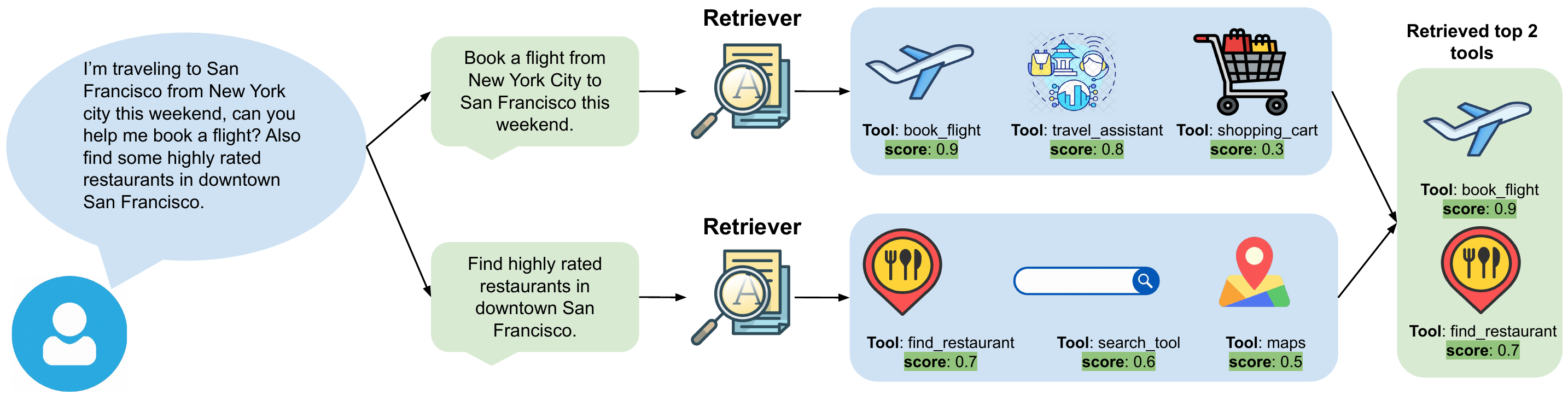}
\caption{An illustration of the multi-view similarity ranking algorithm during retrieval. Multiple intents can be extracted from the user query and we first compute the similarity scores between the expanded tool documents and each intent in the embedding space. We rank and retrieve the top tools from each intent as the final retrieved tools.}
\label{fig:ranking}
\vspace{2mm}
\end{figure*}

\subsection{Query Generator}\label{sec:query_generator}
Tool documents, provided by developers to explain tool usage, are often vague or incomplete, which can lead to incorrect tool retrievals. 
Additionally, the existing text embedding model designed for information retrieval tasks may not accurately model the semantic relationship between tool usage and user queries \citep{patil2023gorilla, qin2023toolllm}.
In practice, tool developers often include usage examples in the tool documents to help users better understanding how to use the tools.

Following this intuition, we instruct LLMs to predict user queries by reading the provided tool document. 
The generated queries then serve as examples of intended tool usages.
We encourage LLMs to produce creative and complex queries that the tool can address.
This process is compatible with any LLMs, including enterprise models such as GPT-4 \citep{gpt4-2023}, Gemini API \citep{gemini-2024}, and open-source models like LLaMa \citep{touvron2023llama}. 
The prompt template is detailed in Appendix~\ref{appendix:prompts}.

To introduce variation and cover the potential query space, we increase the sampling temperature and sample the model response multiple times. 
Examples of the query generator outputs can be found in Appendix~\ref{appendix:response}. 
Finally, each synthesized query is concatenated with the original tool document to create augmented tool documents, facilitating better tool retrievals (see Fig.~\ref{fig:workflow}).

\subsection{Intent Extractor}\label{sec:intent_extractor}
Tool augmented LLM agents often function as chatbots, interacting with users who express their intents in diverse and potentially verbose ways. 
Users may unconsciously provide extraneous background information before stating their actual tasks, which can confuse downstream tool retrieval when trying to identify the underlying intents \citep{qian2024tell}. 
Additionally, users might express multiple intents in a single conversational query, which current retrieval system may struggle to capture due to the query complexity. 
To address these, we leverage LLM's reasoning and query understanding capabilities through in-context learning to extract tool-related intents, thereby improving retrieval accuracy.
This approach also allows for the effective extraction of multiple intents if the user query contains different requests.
We then encode these intents replacing the original user queries into the embedding space during tool retrieval. 
This technique enables the retrieval system to recommend all the relevant tools for each individual intent (see Fig.~\ref{fig:workflow}). 
The prompt template for extracting user intent using LLMs is available in Appendix~\ref{appendix:prompts}.

\subsection{Multi-view Similarity Ranking}
\label{sec:multi-view-ranking}
As each intent extracted from the user queries could retrieve different relevant tools, we introduce a multi-view similarity ranking method to consider all tool-related intents expressed in the user query. 
We aggregate similarity scores between each intent and the expanded tool document. 
By incorporating multiple perspectives within the embedding space, it provides a robust measure of relevance between expanded tool documents and user queries.

We first aggregate the embedding values from multiple copies of the expanded tool document with synthetic queries to represent each tool document in the embedding space.
Instead of grouping the entire tools retrieved by all the intents described in the user query before ranking, we need to rank the tools individually within each intent and retrieve the top tool from each intent until the specified number of tools have been retrieved. 
To achieve this, we design an ordering function to consider both the rank of the retrieved tool within each intent and the similarity score value between the tool and the intent.
The proposed formulation allows us to capture the relevance of each intent to different aspects of the tool document, as represented by the generated queries.

The multi-view similarity ranking algorithm is explained using an example in Fig.~\ref{fig:ranking}. 
The final retrieved 2 tools from the query include the top tool \texttt{book\_flight} retrieved from the intent ``book a flight from New York City to San Francisco this weekend'' and the top tool \texttt{find\_restaurant} retrieved from the intent ``find highly rated restaurants in downtown San Francisco''. The detailed implementation is described in Appendix~\ref{appendix:ranking-algorithm}.


\section{Experimental Settings}
\subsection{Benchmark Datasets}
A variety of benchmark datasets containing tools across different domains have been proposed to assess tool-augmented LLMs. 
These include APIBench \citep{patil2023gorilla}, API-Bank \citep{li2023api}, ToolBench \citep{xu2023tool}, ToolAlpaca \citep{tang2023toolalpaca}, ToolBench \citep{qin2023toolllm}, ToolE \citep{huang2023metatool} and ToolQA \citep{zhuang2023toolqa}. 
To evaluate Re-Invoke's tool retrieval performance, we select ToolBench \citep{qin2023toolllm} and ToolE \citep{huang2023metatool} datasets, as both datasets provide ground truth query and tool document pairs that reflect real-world scenarios. 
We use the same ToolBench dataset to evaluate the end to end performance when integrating the LLM agent with the proposed Re-Invoke retriever. 
Detailed data statistics on the benchmark datasets can be found in Appendix~\ref{appendix:data_statistics}.

\begin{table*}[t!]\small
\centering
\begin{tabular}{l c c | c c c | c c}
    \toprule
    \multirow{2}{*}{\textbf{Retrieval}} &
    \multirow{2}{*}{\textbf{Method}} &
    \multirow{2}{*}{\textbf{Backbone LLM}} &
    \multicolumn{3}{c|}{\textbf{ToolBench}} & \multicolumn{2}{c}{\textbf{ToolE}}\\ 
    \cmidrule{4-8} 
    & & & \textbf{I1} & \textbf{I2} & \textbf{I3} & \textbf{single-tool} & \textbf{multi-tool}  \\
    \midrule
    \multirow{5}{*}{Sparse} & BM25 & - & 0.3475 &  0.2077  & 0.3375 & 0.3735 & 0.2635 \\
     & HyDE w/ Vertex AI & \texttt{text-bison@001} & 0.3084 & 0.1627 & 0.2594 & 0.3770 & 0.1954 \\
     & Re-Invoke w/ BM25 (ours) & \texttt{text-bison@001} & 0.5519 &  0.3968 &  \textbf{0.4990} & 0.5971 & 0.5637  \\
     & Re-Invoke w/ BM25 (ours) & \texttt{gpt-3.5 turbo} & \textbf{0.6013} & \textbf{0.4230} & 0.4959  & \textbf{0.6300} & \textbf{0.5883} \\
     & Re-Invoke w/ BM25 (ours) & \texttt{Mistral-7B} & 0.5768 & 0.3964 & 0.4770  & 0.6134 & 0.5373 \\
    \midrule
    \multirow{5}{*}{Dense} & Vertex AI & -  & 0.5962  & 0.3880  & 0.4633 & 0.6522 & 0.5296 \\
     & HyDE w/ Vertex AI & \texttt{text-bison@001} & 0.4336 & 0.2221 & 0.2996 & 0.6558 & 0.4910 \\
     & Re-Invoke w/ Vertex AI (ours) & \texttt{text-bison@001} & 0.6110  & \textbf{0.5379} & \textbf{0.5955} & \textbf{0.7821} & \textbf{0.7231} \\
     & Re-Invoke w/ Vertex AI (ours) & \texttt{gpt-3.5 turbo} & 0.6090 & 0.5068 & 0.5719  & 0.7705 & 0.6957  \\
     & Re-Invoke w/ Vertex AI (ours) & \texttt{Mistral-7B} & \textbf{0.6150} & 0.5128 & 0.5771  & 0.7770 & 0.6959  \\
    \bottomrule
\end{tabular}
\caption{\label{tab:toolbench-toole-res-ndcg} nDCG@5 metrics on ToolBench I1, I2, I3 and ToolE single-tool, multi-tool datasets using both sparse and retrieval methods. 
In the sparse retrieval method, we apply BM25 retrieval and HyDE retrieval based on BM25 as two baselines. \texttt{text-bison@001} is used to generate hypothesis documents in the HyDE method.
We integrate Re-Invoke with BM25 embedding using \texttt{text-bison@001},  \texttt{gpt-3.5-turbo} and \texttt{Mistral-7B-Instruct-v0.3} as three different backbone LLMs. 
In the dense retrieval method, we apply Vertex AI text embedding retrieval and HyDE retrieval based on Vertex AI text embedding as two baselines. \texttt{text-bison@001} is used to generate hypothesis documents in the HyDE method.
We also integrate Re-Invoke with the Vertex AI text embedding using \texttt{text-bison@001}, \texttt{gpt-3.5-turbo} and \texttt{Mistral-7B-Instruct-v0.3} as three different backbone LLMs. 
The highest nDCG@5 metric is marked in bold.}
\end{table*}

Following the approach in \citet{qin2023toolllm}, we use nDCG@$k$ metric\footnote{\url{https://scikit-learn.org/stable/modules/generated/sklearn.metrics.ndcg_score.html}} to evaluate retrieval performance on the benchmark datasets. 
We report nDCG@5 in the following sections and the detailed retrieval metrics including recall@$k$\footnote{\url{https://www.tensorflow.org/ranking/api_docs/python/tfr/keras/metrics/RecallMetric}} can be found in Appendix~\ref{appendix:retrieval-metrics}. 
For end-to-end performance evaluation, we use pass rate following the same evaluation protocol proposed in \citet{qin2023toolllm}.

\subsection{Unsupervised Retrieval Baselines}
As the proposed method is training free, we establish the following baselines to benchmark Re-Invoke's unsupervised tool retrieval performance: 

\noindent (a) \textbf{Sparse retrieval using BM25}: 
We directly calculate relevance between the user query and the tool documents using BM25. 
We use the default normalization parameter $k=1.5$ for term frequency and offset parameter $b=0.75$ for document length normalization.

\noindent (b) \textbf{Dense retrieval using text embedding}: 
We encode both query and entire tool documents using Google Vertex AI's \texttt{textembedding-gecko@003} model\footnote{\url{https://cloud.google.com/vertex-ai/docs/generative-ai/embeddings/get-text-embeddings}} and compute the cosine similarity on the embedding values.

\noindent (c) \textbf{HyDE as a dense retrieval method}: 
Following \citet{gao2022precise}, we use LLMs to generate a hypothetical tool document for each user query. 
We then calculate document-document similarity using embeddings to retrieve the real tool document. 
In our experiment, we use Google Vertex AI's \texttt{text-bison@001} model\footnote{\url{https://cloud.google.com/vertex-ai/docs/generative-ai/model-reference/text}} to generate the hypothetical tool document. 
The instruction template can be found in Fig.~\ref{fig:hyde} in Appendix~\ref{appendix:prompts}. 
Both hypothetical and real tool documents are encoded using the Vertex AI text embedding API.

\subsection{Re-Invoke}
We use Google Vertex AI's \texttt{text-bison@001} model (with 0.7 temperature) in the query generator to generate 10 diverse synthetic queries per tool document. 
Other parameters remain at default. 
We explored various concatenation methods (prepending, appending, repetition) and found they yield similar retrieval metrics. 
Therefore, we append each generated query to the original tool document in the format: "\texttt{Documentation: <tool document> Query: <predicted query>}" to create different copies of the expanded tool document. 
For evaluation, we vary the number of synthetic queries (see ablation study in Sec.~\ref{sec:ablation}). 

We use the same Google Vertex AI's LLM model in the intent extractor to synthesize the intents from the user queries.
We then extract dense embedding vectors from both augmented tool documents and extracted intents using Google Vertex AI's \texttt{textembedding-gecko@003} model. 
We also apply Re-Invoke using BM25 embedding vectors as a sparse retrieval method.

In our Re-Invoke design, we average the embedding values from multiple copies of the expanded tool document as a representation of the tool document. 
We then compute the embedding similarity score between each extracted intent and the expanded tool documents, and rank the tool documents with the ordering function described in Sec.~\ref{sec:multi-view-ranking}.
We compare our designed aggregation function with others in Sec.~\ref{sec:ablation}.

\section{Experimental Results}
\subsection{Baseline Retrieval Performance}
As shown in Table~\ref{tab:toolbench-toole-res-ndcg}, semantic retrieval using Vertex AI text embedding significantly outperforms the sparse retrieval BM25 across all five benchmark datasets. 
This aligns with the findings in \citet{patil2023gorilla} and \citet{qin2023toolllm}.
Even without specific pre-training on tool retrieval tasks, the existing enterprise text embedding API can effectively represent the semantic relationship between user queries and relevant tool documents.

Compared to its dense retrieval counterpart, HyDE retrieval using Vertex AI text embedding performs less favorably.
This suggests that the HyDE approach introduces a concept drift between the actual and hypothetical tool documents.
The metric degradation can likely be attributed to information loss within the hypothetical documents.

\begin{table*}[t!]\small
\centering
\begin{tabular}{l c c c c c c c}
    \toprule
    \multirow{2}{*}{\textbf{Tool Retriever}} &
    \multicolumn{3}{c}{\textbf{I1 (\%) }} & \multicolumn{2}{c}{\textbf{I2 (\%)}} & \textbf{I3 (\%)} & \multirow{2}{*}{\textbf{Average (\%)}} \\
    \cmidrule{2-7}
    & \textbf{Instruction} & \textbf{Tool} & \textbf{Category} & \textbf{Instruction} & \textbf{Category} & \textbf{Instruction} \\
    \midrule
    None & 39.70 & 44.72 & 47.50 & 64.50 & 55.33 & 61.00 & 52.13 \\
    ToolLLM's & 47.50 & 42.00 & 53.00 & 62.50 & 56.78 & 54.00 & 52.63 \\
    Re-Invoke (ours) & \textbf{48.00} & \textbf{49.75} & \textbf{53.03} & \textbf{65.33} & \textbf{58.29} & \textbf{62.00} & \textbf{56.07} \\
    \bottomrule
\end{tabular}
\caption{\label{tab:toolbench-e2e} End-to-end performance on the ToolBench datasets. 
We follow \citet{qin2023toolllm} to use pass rate as the evaluation metric. 
We integrate ToolLLaMA with DFSDT as the agent, using a set of reference tools without a retriever, ToolLLM's retriever, and Re-Invoke retriever. 
The highest performance metric is marked in bold.}
\vspace{2mm}
\end{table*}

\subsection{Retrieval Performance of Re-Invoke}
Re-Invoke consistently outperforms both sparse and dense retrieval baselines across all benchmark datasets, as shown in Table~\ref{tab:toolbench-toole-res-ndcg} (See Table~\ref{tab:toolbench-toole-all-results} in Appendix for complete results).
When combined with BM25 sparse embeddings, nDCG@5 is significantly increased.
Similarly, Re-Invoke with Vertex AI text embedding yields significant performance gains.
This improvement stems from the proposed LLM-powered tool document enrichment and user intent extraction.

The application of Re-Invoke significantly improves both sparse and dense retrieval performance although we observe that applying Re-Invoke on top of the sparse retrieval method still underperforms the dense retrieval counterparts.
To further analyze the impact of Re-Invoke on improving retrievals, we examine specific user queries and compare the retrieved tools between the baseline and Re-Invoke (see Appendix~\ref{appendix:case-study}). 

We also replicate our experiment using OpenAI's \texttt{gpt-3.5 turbo} model\footnote{\url{https://platform.openai.com/docs/models/gpt-3-5-turbo}} and Mistral AI's \texttt{Mistral-7B-Instruct-v0.3} model \citep{jiang2023mistral} as the backbone LLMs.
The same backbone LLM is used in both the query generator and intent extractor.
Applying the same settings including prompt and decoding parameters as those described in \texttt{text-bison@001}, Re-Invoke achieves a similar trend across all benchmark datasets (see Table~\ref{tab:toolbench-toole-res-ndcg}). 
This demonstrates Re-Invoke's compatibility with various foundation models to improve the baseline retrieval methods.

\subsection{End-to-End Performance Evaluation}
We employ the proposed Re-Invoke as the tool retriever and the ToolLLaMA with Depth-First Search-Based Decision Tree (DFSDT) approach as the agent. 
Please refer to ToolLLM \citep{qin2023toolllm} for comprehensive details on ToolLLaMA with DFSDT implementations. 
We adopt the pass rate metric proposed in ToolLLM \citep{qin2023toolllm} for evaluation metrics. 
Pass rate calculates the percentage of instructions successfully completed within limited budgets. 
We evaluate on six subsets of the ToolBench benchmark dataset: \textit{I1-Instruction}, \textit{I1-Category}, \textit{I1-Tool}, \textit{I2-Instruction}, \textit{I2-Category} and \textit{I3-Instruction}, using OpenAI's \texttt{gpt-3.5-turbo} model as an evaluator.

We compare the agent performance with different tool retriever settings: using a set of reference tools without retrievers, ToolLLM's API retriever \citep{qin2023toolllm} trained using the labeled query-tool pairs, and our Re-Invoke retriever without any training data. 
The reference set of tools are provided in the ToolBench dataset, but they might not be the ground-truth tools as same task could be solved with a different set of tools.
All pass-rate evaluation results are reproduced. 
Table~\ref{tab:toolbench-e2e} demonstrates that our unsupervised Re-Invoke retriever outperforms both baselines with the set of reference tools and a trained tool retriever across all the benchmark datasets. 
This aligns with the finding in ToolLLM \citep{qin2023toolllm} that a tool retriever can expand the search space to find more appropriate tools for a given task. 
Therefore, using tools retrieved by Re-Invoke can improve the agent performance by suggesting more relevant tools given the task even compared to using the reference toolset. 
These evaluation results provide evidence that our Re-Invoke retriever can effectively retrieve relevant tools from a vast pool (16,000+ APIs) and it leads to more reliable downstream agent behaviors on the tool use. 
Importantly, Re-Invoke is completely unsupervised, eliminating the needs for training.

\section{Discussions}
\subsection{Ablation Studies}
\label{sec:ablation}

\begin{table*}[t!] \small
\centering
\begin{tabular}{p{0.4\textwidth}|c|c|c|c|c}
    \toprule
    \multirow{2}{*}{\textbf{Method}} & \multicolumn{3}{c|}{\textbf{ToolBench}} & \multicolumn{2}{c}{\textbf{ToolE}} \\ 
    \cmidrule{2-6}
    & \textbf{I1} & \textbf{I2} & \textbf{I3} & \textbf{single-tool} & \textbf{multi-tool} \\
    \midrule
    \multicolumn{6}{l}{\textsc{(a) Including critical components in Re-Invoke}} \\
    \midrule
    Baseline & 0.5962 & 0.3880 & 0.4633 & 0.6522 & 0.5296 \\
    + Query generator (Sec.~\ref{sec:query_generator}) & \textbf{0.6286} & 0.4135 & 0.4906 & 0.7813 & 0.6906 \\
    + Intent extractor (Sec.~\ref{sec:intent_extractor}) & 0.5910 & 0.5157 & 0.5843 & 0.6756 & 0.6258 \\
    + Query generator \& Intent extractor (Re-Invoke) & 0.6110 & \textbf{0.5379} & \textbf{0.5955} & \textbf{0.7821} & \textbf{0.7231} \\
    \midrule
    \multicolumn{6}{l}{\textsc{(b) Using generated queries only in query generator}} \\
    \midrule
    Synthetic query only & 0.4924	& 0.3050 & 0.4121 & 0.7535 & 0.6814 \\
    Appending the synthetic query to the document & \textbf{0.6286} & \textbf{0.4135} & \textbf{0.4906} & \textbf{0.7813} & \textbf{0.6906} \\
    \midrule
    \multicolumn{6}{l}{\textsc{(c) Varying the number of synthetic queries in query generator}} \\
    \midrule
    1 synthetic query & 0.5962 & 0.3741 & 0.4543 & 0.7388 & 0.6503 \\
    5 synthetic queries & 0.6242 & 0.4091 & 0.4882 & 0.7777 & 0.6724 \\
    10 synthetic queries & \textbf{0.6286} & \textbf{0.4135} & \textbf{0.4906} & \textbf{0.7813} & \textbf{0.6906} \\
     \midrule
    \multicolumn{6}{l}{\textsc{(d) Varying the aggregation function in query generator (Max vs Mean)}} \\
    \midrule
    Maximum similarity score & 0.6104 & 0.3867 & 0.4760 & 0.7716 & 0.6333 \\
    Mean similarity score & \textbf{0.6286} & \textbf{0.4135} & \textbf{0.4906} & \textbf{0.7813} & \textbf{0.6906} \\
    \bottomrule
\end{tabular}
\caption{\label{tab:all_ablation} nDCG@5 metrics of ablation studies on ToolBench I1, I2, I3 datasets and ToolE single-tool, multi-tool datasets. 
We evaluate the impact of each critical component in Re-Invoke: the query generator and the intent extractor. 
Within query generator component, we further compare nDCG@5 metrics across different number of synthetic queries, different aggregation functions to aggregate the relevance scores, and appending the synthetic queries to the tool documents or not. 
The highest retrieval metric is marked in bold.}
\end{table*}

\paragraph{Re-Invoke component evaluation.} 
In this study, we evaluate the tool retrieval performance on each individual Re-Invoke components: query generator and intent extractor using Vertex AI text embedding API. 
The results in Table~\ref{tab:all_ablation}(A) provide the evidence that both designed components contribute positively to final retrieval metrics in Re-Invoke. 
Specifically, we observe consistent retrieval performance improvement across all the benchmark datasets when integrating the query generator with the baseline retrieval method. 
Similar improvement is also demonstrated when applying intent extractor with the baseline. 
When integrating both query generator and intent extractor, Re-Invoke achieves the highest retrieval metrics. 
Note that no improvement is observed on ToolBench I1 dataset when applying intent extractor mainly because ToolBench I1 dataset consists of APIs under the same tool, and each individual intent retrieves overlapped set of APIs. 
We discuss the scenarios when each component performs better in Sec.~\ref{sec:performance-analysis}.

\paragraph{Query generator evaluation.} 
We investigate the retrieval performance with different document augmentation settings using the query generator alone including 
(1) whether to append the document with the synthetic query, 
(2) number of synthetic queries and 
(3) aggregation function. 
Table~\ref{tab:all_ablation}(B,C,D) clearly validate that our design choices in the query generator: appending the synthetic query to the tool document with 10 synthetic queries and aggregate the similarity scores on the augmented tool documents with mean function, outperforms the alternatives. 
We observe that replacing the tool document with the synthetic queries could lead to potential information loss during the retrieval stage and augmenting the tool document is preferred. 
Mean similarity score is more robust when considering diverse synthetic queries in the query generator. 
Increasing the number of diverse synthetic queries improves the retrieval performance and demonstrates the effectiveness in enriching the tool documents with diverse synthetic queries.

\begin{table}[t!]\small
\centering
\begin{tabular}{l | c c c c}
    \toprule
    \multirow{3}{*}{\textbf{Metric}} &
    \multicolumn{3}{c}{\textbf{ToolBench}} & \multicolumn{1}{c}{\textbf{ToolE}} \\
    \cmidrule{2-5}
    & \multirow{2}{*}{\textbf{I1}} & \multirow{2}{*}{\textbf{I2}} & \multirow{2}{*}{\textbf{I3}} & \textbf{single-\&} \\
    &  &  &  & \textbf{multi-tool} \\
    \midrule
    recall@5 & 0.7787 & 0.7665 & 0.9043 & 0.9131 \\
    recall@10 & 0.8402 & 0.8311 & 0.9464 & 0.9462 \\
    \bottomrule
\end{tabular}
\caption{Round-trip consistency recall metrics on synthetic queries in the query generator.}
\label{tab:round-trip-evaluation}
\vspace{2mm}
\end{table}

\subsection{Round-Trip Consistency Evaluation}
We define the synthetic query quality using the round-trip consistency criteria \citep{alberti-etal-2019-synthetic}, \textit{i.e.}, the synthetic queries should retrieve the same tool documents used to generate them.
Specifically, we compute recall@$k$ as a round-trip consistency metric to quantify if the tool document used to generate the synthetic query are among the top-$k$ documents after retrieval using the synthetic query (see Table~\ref{tab:round-trip-evaluation}). 
The relatively high recall@10 metric across all the benchmark datasets suggests query generator's effectiveness to distinguish highly similar tool documents during retrieval.
The round-trip consistency recall metric is lower on ToolBench I1 and I2 datasets, mainly caused by the larger toolset size compared to ToolBench I3 and ToolE datasets.

\subsection{Re-Invoke Performance Analysis} \label{sec:performance-analysis}
We analyze the Re-Invoke performance under different scenarios from the results in Table~\ref{tab:all_ablation}(A).
For relatively short tool documents with minimal human-readable descriptions, tool document expansion alone can significantly boost retrieval performance through generative relevance feedback. 
The documentation in the ToolE datasets only include the tool name and descriptions and we can see that \textit{query generator} alone achieves larger performance gains. 
However, if a tool document lacks API and parameter descriptions, LLMs may struggle to accurately infer usage, relying solely on names. 
This can lead to generated queries that do not reflect real-world tool usage scenarios.
In contrast, for complex user queries with extensive background context or requiring multiple tools simultaneously, \textit{intent extractor} becomes crucial for improving tool retrieval performance.
This component ensures individual tool-related contexts are extracted effectively. 
ToolBench I2 and I3 datasets both contain the queries that need to be handled by calling the APIs from multiple tools and categories on RapidAPI hub, \textit{intent extractor} component alone achieves more significant retrieval performance gains.

\section{Conclusion}
In this work, we present Re-Invoke, a fully unsupervised tool retrieval approach designed to scale LLM tool learning to large toolsets. 
We leverage LLMs to enhance tool document context with diverse synthetic queries, extract essential tool-related intents into executable requests through intent extraction.
Re-Invoke offers a fresh perspective on scalable tool retrieval, prioritizing context enhancement and intent understanding without any training data.

\section{Limitations}

\paragraph{Synthetic query diversity and quality.} 
Re-Invoke achieves query diversity through simple LLM response sampling with a zero-shot prompt. 
To further enhance the quality and reduce concept drift between synthetic and real-world queries, more sophisticated query generation methods could be explored. 
This might include techniques such as controlled prompting, iterative refinement, or utilizing external knowledge bases.

\paragraph{Intent extractor.} 
Re-Invoke relies on in-context learning and LLM's internal knowledge to extract tool related intents. 
Future work could include using the downstream agent's execution results as a feedback to refine intent extraction.

\bibliography{reference}

\appendix
\onecolumn
\section*{Appendix}
\section{Prompt Templates}
\label{appendix:prompts}
In this section, we list all the prompt templates used in our experiments. Fig. \ref{fig:query-generation} is the prompt to generate a synthetic query from the tool document. Fig. \ref{fig:query-rewriting} shows the prompt to extract the user intents. Fig. \ref{fig:hyde} describes the prompt to generate hypothetical tool document given a user query used in the HyDE retrieval baseline.

\begin{figure*}[ht]
\begin{minted}[frame=single,
               framesep=2mm,
               fontsize=\small,
               tabsize=4]{text} 
 
Suppose you are an assistant and you have access to the following API to answer user's queries. 
You are provided with a tool and its available API function including the description and
parameters.

Your task is to generate a possible user query that can be handled by the API.

You must include the input parameters required in the API call. Please be creative and generate 
random but specific information for the required parameters.
Now you are given the API documentation below:

<tool document>

Please generate a user query that you will need to call this tool. Note the generated query should
be complex enough to describe the scenarios that you will need to call the provided API to address
them.

The relevant query is:
\end{minted}
\caption{Prompt template to generate the synthetic queries from the tool document.} \label{fig:query-generation}
\end{figure*}

\newpage
\begin{figure*}[ht]
\begin{minted}[frame=single,
               framesep=2mm,
               fontsize=\small,
               tabsize=4]{text} 
**Instructions**
Suppose you are a query analyzer and your task is to extract the underlying user intents from the
input query. You should preserve all the underlying user request and the extracted user intents 
should be easily understood without extra context information.
You should carefully read the given user query to understand its different intents. Then identify
what are the specific intents. Each individual intent should be separated by a newline.

Here are some examples of how you should solve the task.

**Example**
Query: I'm planning to travel to Paris next weekend to visit my family, could you help me book a 
round trip flight ticket? I want to fly in economy class.
Intent:
book a round-trip flight ticket in economy class to Paris next weekend

Query: I'm a potential buyer looking for a condominium in the city of Miami. I am specifically 
interested in properties that have a minimum of two bathrooms. It should have walkable distance to 
the grocery stores.
Intent:
buy a real estate in Miami with a minimum of two bathrooms and walkable distance to the grocery 
stores

Query: I want to learn Spanish by talking to the native speakers at any time. Additionally, can you
recommend some interesting books, preferably fictions, so that I can learn by reading? Also include
the websites that I can buy them.
Intent:
learn Spanish by talking to the native speakers
recommend fictions to learn Spanish by reading
suggest the websites to buy Sanish fictions

**Begin!**
Query: <user query>
Intent:
\end{minted}
\caption{Prompt template to extract the underlying intents from user queries.} \label{fig:query-rewriting}
\end{figure*}

\newpage
\begin{figure*}[ht!]
\begin{minted}[frame=single,
               framesep=2mm,
               fontsize=\small,
               tabsize=4]{text} 
Suppose you are an assistant and you have access to the API documentation to answer user's queries. 
Please generate an API documentation in the JSON format that can be called to handle this query.
The API documentation should be general enough to handle the cases beyond the provided queries.
Please provide detailed descriptions on the parameters.

**Examples**
Query: I'm planning to travel to Paris next weekend to visit my family, could you help me book a
round trip flight ticket? I want to fly in economy class.
The API documentation is:
{
    "api_name": "flights",
    "api_description": "Search the flight ticket on a specific travel date."
    "required_parameters": [
        {
            "name": "departure_date",
            "type" DATETIME,
            "description": "The departure date for the flight."
        },
        {
            "name": "from",
            "type" STRING,
            "description": "The city where the flight departs."
        },
        {
            "name": "to",
            "type" STRING,
            "description": "The city where the flight arrives."
        },
        {
            "name": "fare_class",
            "type": STRING,
            "description": "The fare class for the flight, economy, business or first."
        }
    ],
    "optional_parameters": [
        {
            "name": "return_date",
            "type": DATETIME,
            "description": "The return date for the flight."
        }
    ]
}

**Begin!**
Query: <user query>
The API documentation is:
\end{minted}
\caption{One-shot prompt template to generate hypothetical tool document given a user query.} \label{fig:hyde}
\end{figure*}

\newpage
\section{Example Synthetic Queries}
\label{appendix:response}

We show 10 different generated user queries from the documentation \texttt{newsSearch} API in the ToolBench dataset in Fig. \ref{fig:example-queries}.

\begin{figure*}[ht!]
\begin{minted}[frame=single,
               framesep=2mm,
               fontsize=\small,
               tabsize=4]{text} 
{"category_name": "Data", "tool_name": "Web Search", "api_name": "newsSearch", "api_description": 
"Get news articles relevant for a given query.", "required_parameters": [{"name": "pageSize", 
"type": "NUMBER", "description": "The number of items per page. The maximum value is 50.", 
"default": "10"}, {"name": "autoCorrect", "type": "BOOLEAN", "description": "Automatically correct
spelling.", "default": true}, {"name": "q", "type": "STRING", "description": "The user's search
query string.", "default": "taylor swift"}, {"name": "pageNumber", "type": "NUMBER", "description": 
"The page to view.", "default": "1"}], "optional_parameters": [{"name"": "toPublishedDate", "type": 
"STRING", "description": "The published date and time for the newest article allowed. For example: 
*2015-05-16T05:50:06.* See  [https://www.c-sharpcorner.com/blogs/date-and-time-format-in-c-sharp
-programming1 ](url)for more possible DateTime formats.", "default": "null"}, {"name": 
"safeSearch", "type": "BOOLEAN", "description": "A filter used to filter results for adult 
content.", "default": false}, {"name": "fromPublishedDate", "type": "STRING", "description": "The 
published date and time for the oldest article allowed. For example:  *2015-05-16T05:50:06.* See 
[https://www.c-sharpcorner.com/blogs/date-and-time-format-in-c-sharp-programming1 ](url)for more 
possible DateTime formats. "", "default": "null"}, {"name": "withThumbnail"", "type": "BOOLEAN", 
"description": "Show results with image thumbnails.", "default": false}], "method": "GET"}
\end{minted}
\begin{minted}[frame=single,
               framesep=2mm,
               fontsize=\small,
               tabsize=4]{text}
Show me news articles about the latest coronavirus outbreak in the United States. I would like to 
see the articles from the past week, and I would like them to be safe for work.

I would like to see a list of news articles about the latest developments in the field of 
artificial intelligence. Please show me the results from the past year and make sure that they are
safe for work.

I'd like to get a list of news articles about the latest developments in the field of artificial 
intelligence. Please show me articles from the past 24 hours, and include images with the results.

I would like to get a list of news articles about the latest developments in artificial 
intelligence. Please show me the results in chronological order, with the most recent articles 
first. I would also like to see images of the articles.

I want to get the news articles about the latest news about the new album by Taylor Swift. I want 
the results to be from the past month and I want them to be safe for work.

I want to know the latest news about the war in Ukraine. Please show me the results from the past 
24 hours, and make sure that they are safe for work.

I want to find news articles about the latest developments in the field of artificial intelligence.
Please show me articles from the past year that are safe for work and have image thumbnails.

I would like to find news articles about the latest developments in AI. Please show me the results
from the past month and make sure to include images in the results.

I would like to know about the latest news articles on the topic of artificial intelligence. 
Please show me the results in chronological order, with the newest articles first. I would also 
like to see thumbnails of the articles.

I would like to find news articles about the latest developments in artificial intelligence. 
Please show me the results from the past week and make sure they are safe for work.
\end{minted}
\caption{The document on the example \texttt{newsSearch} tool and 10 different synthetic queries that are relevant to the provided tool.}
\label{fig:example-queries}
\end{figure*}

\newpage
\section{Multi-view Similarity Ranking Algorithm Implementations}
\label{appendix:ranking-algorithm}

\begin{figure*}[ht]
\centering
\includegraphics[width=0.9\textwidth]{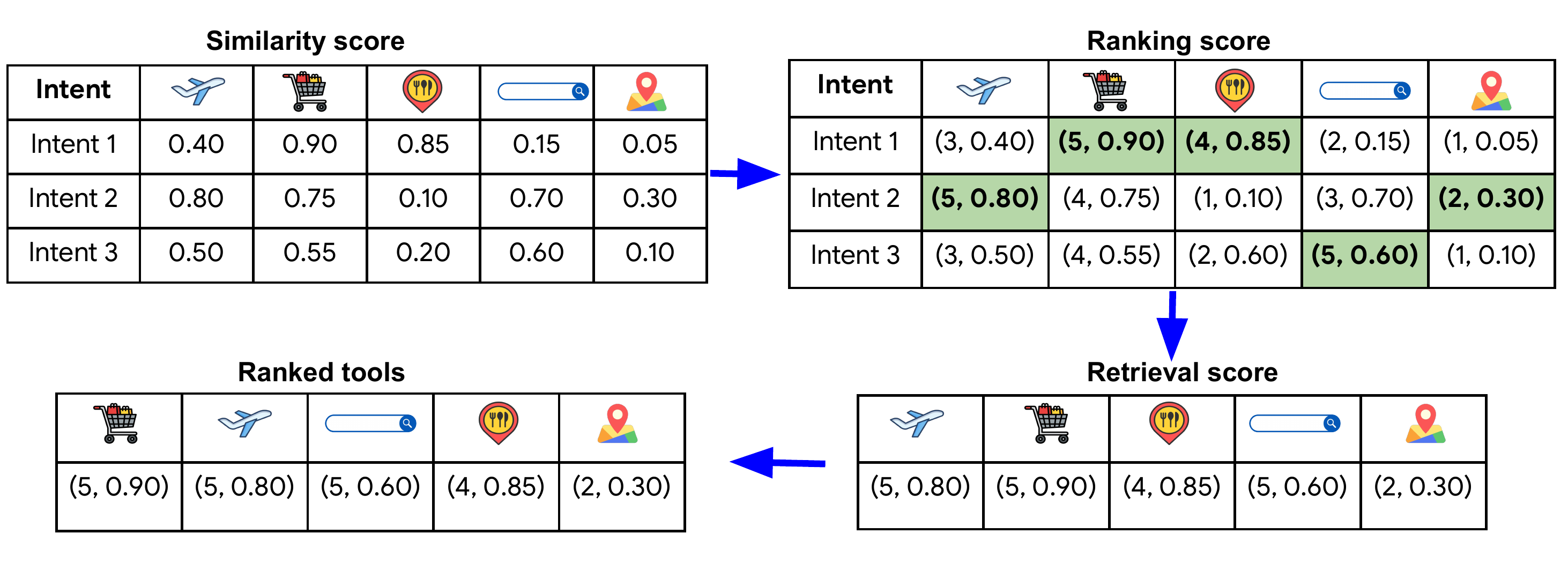}
\caption{An example of the multi-view similarity ranking algorithm. From the intent-tool similarity score, we define the ranking score as a tuple of the reversed ranking order (\textit{i.e.}, lowest similarity score will have a reversed ranking order of 1) for each tool within the same intent and similarity score. We then find the maximum ranking score across multiple intents for each tool document to compute the retrieval score. The retrieval score will be used to retrieve and rank the top tools given the user query.}
\label{fig:ranking-examples}
\vspace{2mm}
\end{figure*}

In this section, we describe the multi-view similarity ranking algorithm (Sec. \ref{sec:multi-view-ranking}) implementations. As each tool document is concatenated with a synthetic query to create $m$ copies, we iterate each augmented tool document and compute the average embedding on each copy of the same tool document. We iterate each user intent from a user query and each tool document to compute the embedding similarity score between the individual intent and tool document. Within each intent, we also compute the reversed ranking order of the tools using the similarity score. To allow each intent to be considered during retrieval, we use a tuple to represent the ranking score to include both the reversed ranking order and similarity score value: we compare the reversed ranking order followed by the similarity score value if the reversed ranking order is the same. We then group all the intents and retrieve the top $k$ documents based on the ranking score. The pseudo-code is available in Algorithm \ref{algo:ranking}, with a detailed working example in Fig. \ref{fig:ranking-examples}.

\begin{algorithm}[!ht]
\caption{Pseudo-code of Re-Invoke's ranking method} \label{algo:ranking}
\KwData{$n$ extracted intents $q_1, q_2, ..., q_n$ from query $Q$, List of tool documents $D$ with each document $d$ concatenated with $m$ synthetic queries and each concatenated document is denoted as $d_1$, $d_2$,..., $d_m$, Text embedding model $f_\text{enc}$}
\KwResult{a retrieval score to rank the documents given a user query}

\SetKwFunction{FRank}{rank}
\SetKwProg{Fn}{Function}{:}{\KwRet {$r(Q, D)$}}
\Fn{\FRank{$q_{1...n}, d_{1...m}, f_\text{enc}$}}{
\For{$d \in D$}{
    $E_d \gets \frac{1}{m} \sum_{i=1}^m (f_\text{enc}(d_i))$\;
}
\For{$i=1, .., n$}
{
    \For{$d \in D$} {
    $\hat{s}(q_i, d) \gets f_\text{enc}(q_i) \cdot E_d$\;
    $\text{rank}(q_i, d) \gets \hat{s}(q_i, d).\texttt{rank}(\texttt{reversed}=\texttt{True}, \texttt{axis}=1)$\;
    $r(q_i, d) \gets (\text{rank}(q_i, d), \hat{s}(q_i, d))$\;
    }
}
$r(Q, D) \gets \texttt{max}_{i=1,2,...,n}r(q_i, D)$\;
}
\end{algorithm}

\newpage
\section{Data statistics}\label{appendix:data_statistics}

The statistics for the benchmark datasets of ToolBench I1, I2, I3 and ToolE single-tool and multi-tool are shown in Table \ref{tab:tool-stats}.

\begin{table*}[ht!] \small
\centering
\begin{tabular}{l c c c}
\toprule
\textbf{Dataset name} & \textbf{Number of queries} & \textbf{Number of tools} & \textbf{Number of labeled pairs} \\
\midrule
ToolBench I1 & 87,419 & 10,439 & 424,169 \\
ToolBench I2 & 84,815 & 13,142 & 220,832 \\
ToolBench I3 & 25,251 & 1,605 & 72,324 \\
\midrule
ToolE single-tool & 20,550 & 199 & 20,550 \\
ToolE multi-tool & 497 & 199 & 994 \\
\bottomrule
\end{tabular}
\caption{\label{tab:tool-stats} Data statistics on ToolBench and ToolE benchmark datasets including number of queries, number of tool documents and number of labeled pairs.}
\end{table*}

\newpage
\section{Retrieval performance evaluation}\label{appendix:retrieval-metrics}

The complete retrieval metrics including nDCG@1, nDCG@5, recall@1 and recall@5 on all the benchmark datasets of ToolBench I1, I2, I3 and ToolE single-tool and multi-tool are shown in Table~\ref{tab:toolbench-toole-all-results}.

\begin{table*}[h!]
\small
\centering
\begin{tabular}{l c c c c c c c}
    \toprule
    \multirow{2}{*}{\textbf{Dataset}} &
    \multirow{2}{*}{\textbf{Retrieval}} &
    \multirow{2}{*}{\textbf{Method}} &
    \multirow{2}{*}{\textbf{LLM}} &
    \multicolumn{2}{c}{\textbf{nDCG}} &
    \multicolumn{2}{c}{\textbf{recall}} \\
    \cmidrule{5-8} & & & & \textbf{@1} & \textbf{@5} &  \textbf{@1} & \textbf{@5} \\
    \midrule
    \multirow{7}{*}{ToolBench I1} & \multirow{5}{*}{Sparse} & BM25 & - & 0.4108 &  0.3588  & 0.1281 & 0.3135 \\
    & & HyDE w/ BM25 & \texttt{text-bison@001} & 0.3449 & 0.3084 & 0.1073 & 0.2729 \\
    &  & Re-Invoke w/ BM25 (ours) & \texttt{text-bison@001} & 0.6338 &  0.5519 &  0.2000 & 0.4767  \\
    &  & Re-Invoke w/ BM25 (ours) & \texttt{gpt-3.5 turbo} & \textbf{0.6809} & \textbf{0.6013} & \textbf{0.2167} & \textbf{0.5225} \\
    &  & Re-Invoke w/ BM25 (ours) & \texttt{Mistral-7B} & 0.6588 & 0.5768 & 0.2089 & 0.4993 \\
    \cmidrule{2-8}
    & \multirow{5}{*}{Dense} & Vertex AI & - & 0.6461 &	 0.5962 & 0.2069	& 0.5278 \\
    &  & HyDE w/ Vertex AI & \texttt{text-bison@001} & 0.4700 & 0.4336 & 0.1508 & 0.3892 \\
    & & Re-Invoke w/ Vertex AI (ours) & \texttt{text-bison@001} & \textbf{0.6947} & 0.6110 & \textbf{0.2231} & 0.5392 \\
    & & Re-Invoke w/ Vertex AI (ours) & \texttt{gpt-3.5 turbo} & 0.6779 & 0.6090 & 0.2209 & 0.5413  \\
    & & Re-Invoke w/ Vertex AI (ours) & \texttt{Mistral-7B} & 0.6847 & \textbf{0.6150} & 0.2207 & \textbf{0.5441}  \\
    \midrule
    \multirow{7}{*}{ToolBench I2} & \multirow{5}{*}{Sparse} & BM25 & - & 0.2543 & 0.2168 & 0.1091 & 0.2201 \\
    & & HyDE w/ BM25 & \texttt{text-bison@001} & 0.1940 & 0.1627 & 0.0839 & 0.1641 \\
    &  & Re-Invoke w/ BM25 (ours) & \texttt{text-bison@001} & 0.4722 & 0.3968 & 0.2007 & 0.3980 \\
    &  & Re-Invoke w/ BM25 (ours) & \texttt{gpt-3.5 turbo} & \textbf{0.4982} & \textbf{0.4230}	& \textbf{0.2117} & \textbf{0.4271} \\
    &  & Re-Invoke w/ BM25 (ours) & \texttt{Mistral-7B} & 0.4691 & 0.3964 & 0.1991 & 0.3990 \\
    \cmidrule{2-8}
    & \multirow{5}{*}{Dense} & Vertex AI & - & 0.4451 & 0.388 & 0.1902 & 0.3976 \\
    &  & HyDE w/ Vertex AI & \texttt{text-bison@001} & 0.2661	& 0.2221 & 0.1144 & 0.2235 \\
    & & Re-Invoke w/ Vertex AI (ours) & \texttt{text-bison@001} & \textbf{0.5456} & \textbf{0.5379}	& \textbf{0.2315}	& \textbf{0.5642} \\
    & & Re-Invoke w/ Vertex AI (ours) & \texttt{gpt-3.5 turbo} & 0.5021 & 0.5068 & 0.2134 & 0.5373  \\
    & & Re-Invoke w/ Vertex AI (ours) & \texttt{Mistral-7B} & 0.5133 & 0.5128 & 0.2174 & 0.5426  \\
    \midrule
    \multirow{7}{*}{ToolBench I3} & \multirow{5}{*}{Sparse} & BM25 & - & 0.4075	& 0.3452 & 0.1601 & 0.3480 \\
    &  & HyDE w/ BM25 & \texttt{text-bison@001} & 0.2965 & 0.2594 & 0.1162 & 0.2655 \\
    &  & Re-Invoke w/ BM25 (ours) & \texttt{text-bison@001} & \textbf{0.5768} & \textbf{0.4990} & \textbf{0.2253} & \textbf{0.5052}  \\
    &  & Re-Invoke w/ BM25 (ours) & \texttt{gpt-3.5 turbo} & 0.5660 & 0.4959	& 0.2213 & 0.5052 \\
    &  & Re-Invoke w/ BM25 (ours) & \texttt{Mistral-7B} & 0.5432 & 0.4770 & 0.2124 & 0.4868 \\
    \cmidrule{2-8}
    & \multirow{5}{*}{Dense} & Vertex AI & - & 0.5165 & 0.4633 & 0.2029	& 0.4793 \\
    &  & HyDE w/ Vertex AI & \texttt{text-bison@001} & 0.3475	& 0.2996 & 0.1376 & 0.3048 \\
    & & Re-Invoke w/ Vertex AI (ours) & \texttt{text-bison@001} & \textbf{0.5965} & \textbf{0.5955}	& \textbf{0.2327}	& \textbf{0.6242} \\
    & & Re-Invoke w/ Vertex AI (ours) & \texttt{gpt-3.5 turbo} & 0.5764 & 0.5719 & 0.2248 & 0.6016  \\
    & & Re-Invoke w/ Vertex AI (ours) & \texttt{Mistral-7B} & 0.5833 & 0.5771 & 0.2271 & 0.6047  \\
    \midrule
    \multirow{7}{*}{ToolE single-tool} & \multirow{5}{*}{Sparse} & BM25 & - & 0.2716 & 0.3732 & 0.2716 & 0.4618 \\
    &  & HyDE w/ BM25 & \texttt{text-bison@001} & 0.3121 & 0.3770 & 0.3121 & 0.4337 \\
    &  & Re-Invoke w/ BM25 (ours) & \texttt{text-bison@001} & 0.4928 & 0.5971 & 0.4927 & 0.6866 \\
    &  & Re-Invoke w/ BM25 (ours) & \texttt{gpt-3.5 turbo} & \textbf{0.5255} & \textbf{0.6300} & \textbf{0.5255}	& \textbf{0.7193} \\
    &  & Re-Invoke w/ BM25 (ours) & \texttt{Mistral-7B} & 0.5021 & 0.6134 & 0.5021 & 0.7093 \\
    \cmidrule{2-8}
    & \multirow{5}{*}{Dense} & Vertex AI & - & 0.5265 & 0.6522 & 0.5265 & 0.7574 \\
    &  & HyDE w/ Vertex AI & \texttt{text-bison@001} & 0.5488	& 0.6558 & 0.5488 & 0.7448 \\
    & & Re-Invoke w/ Vertex AI (ours) & \texttt{text-bison@001} & \textbf{0.6716} & \textbf{0.7821}	& \textbf{0.6715}	& \textbf{0.8707} \\
    & & Re-Invoke w/ Vertex AI (ours) & \texttt{gpt-3.5 turbo} & 0.6551 & 0.7705 & 0.6551 & 0.8635  \\
    & & Re-Invoke w/ Vertex AI (ours) & \texttt{Mistral-7B} & 0.6600 & 0.7770 & 0.6600 & 0.8714  \\
    \midrule
    \multirow{7}{*}{ToolE multi-tool} & \multirow{5}{*}{Sparse} & BM25 & - & 0.1841	& 0.2627 & 0.0926 & 0.3350 \\
    &  & HyDE w/ BM25 & \texttt{text-bison@001} & 0.2414 & 0.1954 & 0.1207 & 0.1942 \\
    &  & Re-Invoke w/ BM25 (ours) & \texttt{text-bison@001} & 0.5392 & 0.5637 & 0.2696 & 0.6408  \\
    &  & Re-Invoke w/ BM25 (ours) & \texttt{gpt-3.5 turbo} & \textbf{0.5634} & \textbf{0.5883} & \textbf{0.2817}	& \textbf{0.6610} \\
    &  & Re-Invoke w/ BM25 (ours) & \texttt{Mistral-7B} & 0.4909 & 0.5373 & 0.2455 & 0.6157 \\
    \cmidrule{2-8}
    & \multirow{5}{*}{Dense} & Vertex AI & - & 0.4286 & 0.5296 & 0.2143 & 0.6258 \\
    &  & HyDE w/ Vertex AI & \texttt{text-bison@001} & 0.5614 & 0.4910	& 0.2807 & 0.5050 \\
    & & Re-Invoke w/ Vertex AI (ours) & \texttt{text-bison@001} & \textbf{0.6660} & \textbf{0.7231} &	\textbf{0.3330}	& \textbf{0.8008} \\
    & & Re-Invoke w/ Vertex AI (ours) & \texttt{gpt-3.5 turbo} & 0.6499 & 0.6957 & 0.3249	& 0.7797  \\
    & & Re-Invoke w/ Vertex AI (ours) & \texttt{Mistral-7B} & 0.6076 & 0.6959 & 0.3038 & 0.7968  \\
\bottomrule
\end{tabular}
\caption{\label{tab:toolbench-toole-all-results} Retrieval metrics (nDCG@1, nDCG@5, recall@1 and recall@5) on ToolBench I1, I2, I3 and ToolE single-tool and multi-tool datasets with different approaches including baselines and Re-Invoke using BM25 and Vertex AI text embedding. We observe the similar tool retrieval performance trend with different retrieval metrics. The highest metric is marked in bold.}
\end{table*}

\newpage
\section{Case Study on Retrieved Tools}
\label{appendix:case-study}

In this section, we showcase that tool retrieval can benefit from Re-Invoke using the ToolE dataset as demonstration examples. A few user queries are cherrypicked from the ToolE single-tool and multi-tool datasets that the baseline Vertex AI retriever retrieved the wrong tools while the Re-Invoke recommended the relevant tools. We investigate query generator and intent extractor components separately.

Table \ref{tab:case-study-correct-expansion} shows the example queries that the correct tools are retrieved with the query generator component. It can be clearly seen that Re-Invoke's query generator component can better distinguish among similar tools to determine which tool is more relevant to user's request. For example, when the user is asking for the weather forecast for a location, the Vertex AI baseline retriever retrieves the very specific \texttt{airqualityforeast} tool while Re-Invoke retrieves the correct \texttt{WeatherTool} tool, which is more tailed to answer user's queries.

Table \ref{tab:case-study-correct-intent} lists the correct tools retrieved with the intent extractor component.  Similarly, Re-Invoke's intent extractor effectively understands the user intents to recommend the most relevant tools to user's specific request. For example, when the user is asking for recommendations on online courses on machine learning and needs the access to relevant PDFs or URLs, Re-Invoke's intent extractor identifies two intents ``recommend a course on machine learning'' and ``have access to relevant PDFs or URLs for further reading'' and successfully retrieves the correct tools \texttt{CourseTool} and \texttt{PDF\&URLTool} from each intent. However, the baseline retrieval method retrieves \texttt{CourseTool} and \texttt{search} tools instead.

We have also observed that Re-Invoke can still lead to wrong retrievals, especially when the tools are very similar, e.g., \texttt{HousePurchasingTool} and \texttt{HouseRentingToo}, \texttt{FinanceTool} and \texttt{CompanyInfoTool}. Please see the examples in Table \ref{tab:case-study-incorrect}. When the user is explicitly looking to buy a condominium in the query, Re-Invoke retrieves the wrong \texttt{HouseRentingTool}. We believe those errors can be reduced by designing a more sophisticated approach to generate more explicit synthetic queries that can be used to distinguish among confusing tool documents. 

\newpage
\begin{table*}[ht] \small
    \centering
    \begin{tabular}{p{0.2\linewidth} p{0.7\linewidth} }
    \toprule
    \textbf{User query} & Are there any strategy games, specifically turn-based and real-time strategy games, available for Nintendo Switch that offer a variety of gameplay options and customizable gameplay mechanics? \\
    \midrule
    \textbf{Baseline retrievals} & \hlpink{\{"name": "Chess", "description": "Unleash your inner chess master with this interactive chess experience! You can play against a novice or a grandmas-ter!"\}} \\
    \midrule
    \textbf{Re-Invoke retrievals} & \hllime{\{ "name": "GameTool", "description":"Get game-related information and recommend games."\}} \\
    \bottomrule
    \toprule
    \textbf{User query} & What  are  some  popular  attractions in London that I  shouldn't miss? I’m specifically interested in historical landmarks, famous museums, and iconic landmarks that are unique to the city. \\
    \midrule
    \textbf{Baseline retrievals} & \hlpink{\{"name":"themeparkhipster", "description": "Find theme park waiting times around the world."\}} \\
    \midrule
    \textbf{Re-Invoke retrievals} & \hllime{\{ "name": "TripAdviceTool", "description": "A comprehensive travel assistant that makes travel planning more vivid and practical. It offers tourism activities, accommodation and attraction recommendations, aiming to provide users with a more enjoyable and enriching travel experience through technology."\}} \\
    \bottomrule
    \toprule
    \textbf{User query} & Can you help me find research papers on a specific topic? \\
    \midrule
    \textbf{Baseline retrievals} & \hlpink{\{"name": "ResearchHelper", "description": "Tool that offers additional functions beyond searching academic papers, such as generating mind maps, answering user questions and storing them in specific formats."\}} \\
    \midrule
    \textbf{Re-Invoke retrievals} & \hllime{\{ "name": "ResearchFinder", "description": "Tool for searching academic papers."\}} \\
    \bottomrule
    \toprule
    \textbf{User query} & Could you please provide me with the highly detailed weather forecast for Tokyo, the capital city of Japan, specifically for the upcoming four days? \\
    \midrule
    \textbf{Baseline retrievals} &  \hlpink{\{"name": "airqualityforeast", "description": "Planning something outdoors? Get the 2-day air quality forecast for any US zip code."\}} \\
    \midrule
    \textbf{Re-Invoke retrievals} & \hllime{\{ "name": "WeatherTool", "description": "Provide you with the latest weather information."\}} \\
    \bottomrule
    \toprule
    \textbf{User query} & Could you please provide detailed information about the deployed smart contract of a specific ERC20 token on the Ethereum blockchain, including its address, source code, contract functions, and any associated events or transactions? \\
    \midrule
    \textbf{Baseline retrievals} & \hlpink{\{"name": "magi\_codex", "description": "Ask about Magic: The Gathering cards, rules and interactions."\}} \\
    \midrule
    \textbf{Re-Invoke retrievals} & \hllime{\{"name": "FinanceTool", "description": "Stay informed with the latest financial updates, real-time insights, and analysis on a wide range of options, stocks, cryptocurrencies, and more."\}} \\
    \bottomrule
    \toprule
    \textbf{User query} & What questions can I ask about this YouTube video? \\
    \midrule
    \textbf{Baseline retrievals} & \hlpink{\{"name": "video\_highlight", "description": "Explore, research, and interact with YouTube videos and personal videos."\}} \\
    \midrule
    \textbf{Re-Invoke retrievals} & \hllime{\{"name": "VideoSummarizeTool", "description": "Generate summaries from YouTube video links, offer question-answering capabilities, analyze and interpret the content of YouTube videos, and support interactions with online video platforms such as YouTube and Daily Motion."\}} \\
    \bottomrule
    \toprule
    \textbf{User query} & Despite submitting my resume to numerous companies and job openings, I have not received any responses or feedback regarding my application. \\
    \midrule
    \textbf{Baseline retrievals} & \hlpink{\{"name": "JobTool", "description": "Your Global Career Hub! Find diverse job opportunities, expert interview tips, and resume optimization guidance. Empowering job seekers worldwide on their path to success."\}} \\
    \midrule
    \textbf{Re-Invoke retrievals} & \hllime{\{"name": "ResumeTool", "description": "Quickly create resumes and receive feedback on your resume."\}} \\
    \bottomrule
    \end{tabular}
    \caption{A list of cherry-picked example queries from the ToolE single-tool dataset, including top 1 tool retrieved by the baseline and Re-Invoke's query generator using the Vertex text embedding API. Re-Invoke's query generator retrieves the correct tools (in \hllime{green}) while the baseline retrieves the wrong tools (in \hlpink{red}).}
    \label{tab:case-study-correct-expansion}
\end{table*}

\newpage
\begin{table*}[ht] \small
    \centering
    \begin{tabular}{p{0.2\linewidth} p{0.7\linewidth} }
    \toprule
    \textbf{User query} & What are \hllime{some popular investment options with good returns}, and \hlcyan{can you recommend a playlist to relax} while I research them? \\
    \midrule
    \textbf{Baseline retrievals} & \hllime{\{"name": "FinanceTool", "description": "Stay informed with the latest financial updates, real-time insights, and analysis on a wide range of options, stocks, cryptocurrencies, and more."\}} \\
    & \hlpink{\{"name": "portfoliopilot", "description": "Your AI investing guide: portfolio assessment, recommendations, answers to all finance questions."\}} \\
    \midrule
    \textbf{Re-Invoke retrievals} & \hllime{\{"name": "FinanceTool", "description": "Stay informed with the latest financial updates, real-time insights, and analysis on a wide range of options, stocks, cryptocurrencies, and more."\}} \\
    & \hlcyan{\{"name": "MusicTool", "description": "Create music playlists, search for music, and check out the latest music trends."\}} \\
    \bottomrule
    \toprule
    \textbf{User query} & Could you provide me with \hllime{news articles on renewable energy sources} and any \hlcyan{research papers exploring their effectiveness?}? \\
    \midrule
    \textbf{Baseline retrievals} & \hlcyan{\{"name": "ResearchHelper", "description": "Tool that offers additional functions beyond searching academic papers, such as generating mind maps, answering user questions and storing them in specific formats."\}} \\
    & \hlpink{\{"name": "ResearchFinder", "description": "Tool for searching academic papers."\}} \\
    \midrule
    \textbf{Re-Invoke retrievals} & \hlcyan{\{"name": "ResearchHelper", "description": "Tool that offers additional functions beyond searching academic papers, such as generating mind maps, answering user questions and storing them in specific formats."\}} \\
    & \hllime{\{"name": "NewsTool", "description": "Stay connected to global events with our up-to-date news around the world."\}} \\
    \bottomrule
    \toprule
    \textbf{User query} & Can you \hllime{recommend me a course on machine learning}? I want to learn more about the topic, and also \hlcyan{have access to relevant PDFs or URLs for further reading}. \\
    \midrule
    \textbf{Baseline retrievals} & \hllime{\{"name": "CourseTool", "description": "Unlock a world of knowledge and growth with our comprehensive learning platform, offering a diverse range of courses from renowned providers like Coursera and Upskillr, personalized language learning, professional team information lookup, open course schedule discovery, and top-tier university content."\}} \\
    & \hlpink{\{"name": "search", "description": "Level up your design skills quickly with a wide range of design courses, interactive workshops and AI-guided mentorship."\}} \\
    \midrule
    \textbf{Re-Invoke retrievals} & \hlcyan{\{"name": "PDF\&URLTool", "description": "Interact with any PDF files, provide page references for fact-checking, support chatting via Google Drive links to AI-driven PDF summaries and analysis; engage in interactive conversations with websites, access links on the internet to fetch required information, including generating articles and intelligent assistance for interactions with source code."\}} \\
    & \hllime{\{"name": "CourseTool", "description": "Unlock a world of knowledge and growth with our comprehensive learning platform, offering a diverse range of courses from renowned providers like Coursera and Upskillr, personalized language learning, professional team information lookup, open course schedule discovery, and top-tier university content."\}} \\
    \bottomrule
    \toprule
    \textbf{User query} & Can you \hllime{recommend any online courses for learning about natural language processing} and \hlcyan{a GitHub repository with relevant code examples}? \\
    \midrule
    \textbf{Baseline retrievals} & \hllime{\{"name": "CourseTool", "description": "Unlock a world of knowledge and growth with our comprehensive learning platform, offering a diverse range of courses from renowned providers like Coursera and Upskillr, personalized language learning, professional team information lookup, open course schedule discovery, and top-tier university content."\}} \\
    & \hlpink{\{"name": "AI2sql", "description": "Converts a natural language text into an SQL query."\}} \\
    \midrule
    \textbf{Re-Invoke retrievals} & \hlcyan{\{"name": "RepoTool", "description": "Discover GitHub projects tailored to your needs, explore their structures with insightful summaries, and get quick coding solutions with curated snippets. Elevate your coding journey with RepoTool, your go-to companion for GitHub project exploration and code mastery."\}} \\
    & \hllime{\{"name": "CourseTool", "description": "Unlock a world of knowledge and growth with our comprehensive learning platform, offering a diverse range of courses from renowned providers like Coursera and Upskillr, personalized language learning, professional team information lookup, open course schedule discovery, and top-tier university content."\}} \\
    \bottomrule
    \end{tabular}
    \caption{A list of cherry-picked example queries from the ToolE multi-tool dataset, including 2 tools retrieved by the baseline and Re-Invoke's intent extractor component using the Vertex text embedding API. Re-Invoke's intent extractor identifies the intents (in \hllime{green} and \hlcyan{blue}) and retrieves the correct tools (in \hllime{green} and \hlcyan{blue}) while the baseline retrieves the wrong tools (in \hlpink{red}).}
    \label{tab:case-study-correct-intent}
    \vspace{-8mm}
\end{table*}

\newpage
\begin{table*}[ht] \small
    \centering
    \begin{tabular}{p{0.2\linewidth} p{0.7\linewidth} }
    \toprule
    \textbf{User query} & I'm working on a project about renewable energy sources. Can you help me find scientific publications related to this topic and generate bibtex bibliographies for them? \\
    \midrule
    \textbf{Baseline retrievals} & \hllime{\{"name": "ResearchHelper", "description": "Tool that offers additional functions beyond searching academic papers, such as generating mind maps, answering user questions and storing them in specific formats."\}} \\
    \midrule
    \textbf{Re-Invoke retrievals} & \hlpink{\{ "name": "ResearchFinder", "description": "Tool for searching academic papers."\}} \\
    \bottomrule
    \toprule
    \textbf{User query} & I'm a potential buyer looking for a condominium in the city of Miami. I am specifically interested in properties that have a minimum of two bathrooms. \\
    \midrule
    \textbf{Baseline retrievals} & \hllime{\{"name": "HousePurchasingTool", "description": "Tool that provide all sorts of information about house purchasing"\}} \\
    \midrule
    \textbf{Re-Invoke retrievals} & \hlpink{\{ "name": "HouseRentingTool", "description": "Tool that provide all sorts of information about house renting"\}} \\
    \bottomrule
    \toprule
    \textbf{User query} & Can you provide me with a list of restaurants in Japan that I can reserve a table at? \\
    \midrule
    \textbf{Baseline retrievals} & \hllime{\{"name": "RestaurantBookingTool", "description": "Tool for booking restaurant"\}} \\
    \midrule
    \textbf{Re-Invoke retrievals} & \hlpink{\{ "name": "TripAdviceTool", "description": "A comprehensive travel assistantn that makes travel planning more vivid and practical. It offers tourism activities, accommodation and attraction recommendations, aiming to provide users with a more enjoyable and enriching travel experience through technology." \}} \\
    \bottomrule
    \toprule
    \textbf{User query} & I'm looking for a luxurious hotel in Bali that offers stunning beach views, as I am planning a romantic getaway. \\
    \midrule
    \textbf{Baseline retrievals} & \hllime{\{"name": "TripTool", "description": "Offer discounted hotel and accommodation bookings, along with personalized hotel and product searches, travel planning, image editing, and more, helping users easily plan their trips and find accommodation and transportation options."\}} \\
    \midrule
    \textbf{Re-Invoke retrievals} & \hlpink{\{ "name": "TripAdviceTool", "description": "A comprehensive travel assistantn that makes travel planning more vivid and practical. It offers tourism activities, accommodation and attraction recommendations, aiming to provide users with a more enjoyable and enriching travel experience through technology."\}} \\
    \bottomrule
    \toprule
    \textbf{User query} & Users would like to know the exact percentage of the dividend yield for Coca-Cola stock based on its current market price and dividend payouts. \\
    \midrule
    \textbf{Baseline retrievals} & \hllime{\{"name": "FinanceTool", "description": "Stay informed with the latest financial updates, real-time insights, and analysis on a wide range of options, stocks, cryptocurrencies, and more."\}} \\
    \midrule
    \textbf{Re-Invoke retrievals} & \hlpink{\{"name": "CompanyInfoTool", "description": "Obtain relevant information about global companies from databases or knowledge graphs."\}} \\
    \bottomrule
    \toprule
    \textbf{User query} & I have a presentation scheduled in the near future, and I am seeking assistance in enhancing its impact. Specifically, I would greatly appreciate it if you could utilize your skills to paraphrase and rephrase significant elements extracted from my research article, thereby transforming them into highly captivating and attention-grabbing points. \\
    \midrule
    \textbf{Baseline retrievals} & \hllime{\{"name": "SummarizeAnything\_pr", "description": "Summarize YouTube videos, web pages, and PDF documents by providing a link. This is a free preview."\}} \\
    \midrule
    \textbf{Re-Invoke retrievals} & \hlpink{\{"name": "PolishTool", "description": "Elevate your content with our AI-powered tool, which utilizes advanced rewriting techniques to create more human-like expressions and foster creative inspiration."\}} \\
    \bottomrule
    \end{tabular}
    \caption{A list of cherrypicked example queries from the ToolE single-tool dataset, including top 1 tool retrieved by the baseline and Re-Invoke using the Vertex text embedding API. Baseline retrieves the correct tool (in \hllime{green}), while Re-Invoke retrieves the wrong tool (in \hlpink{red}).}
    \label{tab:case-study-incorrect}
\end{table*}

\end{document}